\documentclass[11pt]{article}
\usepackage[final]{acl}

\usepackage{times}
\usepackage{latexsym}
\usepackage[most]{tcolorbox}
\usepackage[T1]{fontenc}

\usepackage[utf8]{inputenc}
\usepackage[colorinlistoftodos,prependcaption,textsize=tiny]{todonotes}
\usepackage{microtype}
\usepackage{enumitem}
\usepackage{inconsolata}

\usepackage{graphicx}
\usepackage{amsmath}
\usepackage{makecell}
\usepackage{multirow}
\usepackage{booktabs}
\usepackage{threeparttable}
\usepackage{xcolor}

\title{How Do Lexical Senses Correspond Between \\Spoken German and German Sign Language?}

\author{Melis Çelikkol\textsuperscript{1}\, and \,
Wei Zhao\textsuperscript{2} \\[0.4em]
  Institute for Computational Linguistics, University of Heidelberg\textsuperscript{1} \\
  Department of Computing Science, University of Aberdeen\textsuperscript{2}\\
  \href{mailto:melis.celikkol@stud.uni-heidelberg.de}{melis.celikkol@stud.uni-heidelberg.de}\\ 
  \href{mailto:wei.zhao@abdn.ac.uk}{wei.zhao@abdn.ac.uk}
}

\begin{document}
\maketitle
\begin{abstract}

Sign language lexicographers construct bilingual dictionaries by establishing word-to-sign mappings, where polysemous and homonymous words corresponding to different signs across contexts are often underrepresented. A usage-based approach examining how word senses map to signs can identify such novel mappings absent from current dictionaries, enriching lexicographic resources.
We address this by analyzing German and German Sign Language (Deutsche Gebärdensprache, DGS), manually annotating 1,404 word use–to–sign ID mappings 
derived from 32 words from the German Word Usage Graph (D-WUG) and 49 signs from the Digital Dictionary of German Sign Language (DW-DGS). We identify three correspondence types: Type 1 (one-to-many), Type 2 (many-to-one), and Type 3 (one-to-one), plus No Match cases. We evaluate computational methods: Exact Match (EM) and Semantic Similarity (SS) using SBERT embeddings. SS substantially outperforms EM overall 88.52\% vs. 71.31\%), with dramatic gains for Type 1 (+52.1 pp). Our work establishes the first annotated dataset for cross-modal sense correspondence and reveals which correspondence patterns are computationally identifiable.
Our code and dataset are made publicly available\footnote{\url{https://github.com/C-Melis/Ambiguity_Resolution_Across_German_Words_and_Their_Sign_Correspondents}}.
\end{abstract}

\section{Introduction}

Sign language lexicographers construct bilingual dictionaries by establishing word-to-sign mappings, typically documenting one canonical mapping per word. However, polysemous and homonymous words often correspond to multiple distinct signs across different contexts, yet existing dictionaries may not capture this full range of correspondences. A usage-based approach that examines how word senses map to signs across word usages can identify such novel mappings absent from current dictionaries, enriching lexicographic resources and revealing systematic patterns in how ambiguities transfer across the two modalities.

Lexical ambiguity arises when the meaning of a word changes across different contexts, making its actual sense uncertain until the context is specified. This uncertainty of word sense exists in all languages. Even when two languages use the "same" word,
their senses do not align one-to-one. For instance, English \textit{bank} refers to a financial institution or the side of a river, while German \textit{Bank} does not cover the sense of river side. This shows that languages often differ in how senses are encoded within a word and its translation. Identifying such sense correspondence between word translations 
is crucial in lexicography, language learning and computational linguistics \citep{hurford2007semantics,simatupang2007ambiguous}, as 
this will help lexicographers to build dictionaries.

Identifying sense correspondence becomes more challenging when we compare between spoken and sign languages. 
Sign languages, as fully developed natural languages operating in the visual-gestural modality, 
also exhibit lexical ambiguity, where the senses of a sign may align with, deviate from, or partially overlap with the senses of its word translation in a spoken language. This leads to unparallel senses between a word and its sign translation(s) 
\citep{johnston2007australian,quer2015ambiguities}. 
Characterizing these correspondence patterns empirically, identifying which patterns exist, and whether computational systems can reliably detect them, remains an open question. 
For instance, the German word "erlauben" (allow/permit) has three sign translations in DGS, where multiple senses are encoded within a single word form,
whereas DGS distributes these senses across three signs. This shows that senses correspond differently across the spoken and sign modalities. 
 
While sense correspondence (one-to-many, many-to-one, and one-to-one) across spoken languages has been investigated \citep{xu-etal-2024-annotating, rahit-etal-2018-banglanet}, little attention has been paid to sense correspondence between spoken and sign languages, as available resources lack semantic annotations required to compare sense correspondence.  
Although previous studies showed that semantic differences are present across the two modalities \citep{schulder2024signs}, existing methods cannot identify how these differences exhibit, especially whether polysemy and homonymy in spoken languages mirror, diverge from, or partially overlap with those in sign languages.

In this project, our aim is to identify the types of sense correspondence between spoken German and German Sign Language (Deutsche Gebärdensprache DGS). To do so, we manually annotate words and their sign
correspondence based on two
linguistic 
resources: (i) the German Word Usage Graph (D-WUG) \citep{schlechtweg-etal-2024-refwug}, 
providing word uses, and (ii) the Digital Dictionary of German Sign Language 
(DW-DGS) \citep{langer-etal-2024-introducing}, containing signs through video recordings 
with unique identifier labels, German translations, and ``Erklärung'' (the explanation of the sense, the so-called dictionary definition). All words 
selected from D-WUGs exhibit multiple senses, making them inherently ambiguous and diversifying the types of  sense correspondence across the two modalities. By matching German words from 
D-WUG to their sign translations in DW-DGS, we create a manually-annotated 
dataset containing three types of cross-modal sense correspondence. 
Our work makes three key contributions to computational sign language research:

\begin{itemize}
    \item 
    We provide 1,404 human-annotated mappings from word uses to sign IDs  (973 train+val+test\_overlap uses, 431 test\_no\_overlap uses) derived from 32 German words, establishing the first resource for analyzing cross-modal ambiguity correspondence grounded in word usages.
    \item 
    We identify and characterize three distinct patterns of how ambiguities transfer across modalities: Type 1 (one-to-many, 28.6\% of words), Type 2 (many-to-one, 28.6\%), and Type 3 (one-to-one, 33.3\%), demonstrating that no single pattern dominates cross-modal semantic organization.
    \item 
    Our semantic similarity method achieves 88.52\% accuracy overall, with dramatic improvements for Type 1 (+52.1 pp over exact matching), revealing which types of correspondence are easy to identify and which are not.

\end{itemize}

\section{Related Work}

\paragraph{Lexical Ambiguities}
result from the fact that a single word form can have multiple meanings,
primarily through polysemy and homonymy \citep{klepousniotou2002processing,haber2024polysemy}. Polysemy associates one word with multiple conceptually or historically related senses sharing a common etymological core, while homonymy involves words sharing identical form but having unrelated senses \citep{fromkin2010introduction}.
Similarly, sign languages exhibit lexical ambiguities, with many signs being ambiguous or multifunctional \citep{quer2015ambiguities,pfau2016modality}. This is due to its distinction from spoken languages:
 modality-specific properties of sign languages, such as the three-dimensional signing space for establishing referential loci,  the simultaneous use of manual and non-manual articulators, role shift enabling perspective adoption, and spatial modification of classifier hand shapes, actively affect their linguistic structure \citep{johnston2007australian}. Previous work shows that polysemy and homonymy are present across various
 sign languages \citep{gwammaja2025semantic,neubauer2024popsign}, demonstrating that sign languages are living languages harboring these phenomena actively \citep{bahan1996come}.

To disambiguate the senses of a word or a sign, different disambiguation approaches are applied:
For spoken languages, previous work relies on context based on surrounding words and discourse information to resolve lexical ambiguities \citep{fromkin2010introduction}. 
For sign languages, previous work employs modality-specific methods (gestures, spatial modification of classifier hand shapes, non-manual markers), cross-modal methods (mouthing), and modality-independent methods (such as context,
and anaphora resolution) \citep{johnston2007australian,quer2015ambiguities,grimm2024signlanguagesensedisambiguation}.

\paragraph{Computational Approaches to Sign Languages.}
Despite progress in sign recognition 
\citep{alabdullah2024advancements}, most work focuses on isolated signs (recognising one sign at a time) 
rather than 
a sequence of signs, 
covering not more than 50 signs 
\citep{koller2020quantitative,alabdullah2024advancements}. Recent 
work focuses on creating new datasets:
the PopSign ASL 
dataset \citep{starner2023popsign} enables recognition of isolated signs, 
\citet{neubauer2024popsign} identifies confusion patterns among visually similar signs 
(addressing homonym identification), and \citet{ortega2025lexical} document iconicity 
and concreteness norms across BSL and DGS.
\citet{grimm2024signlanguagesensedisambiguation} address computational disambiguation for sign languages by using transformer-based models on the RWTH-PHOENIX-Weather Database \citep{koller2015continuous}, finding that approximately 64\% of cases represent homonymous expressions but noting that fine-grained semantic disambiguation remains challenging. \citet{schulder2024signs} develop the Multilingual Sign Language Wordnet (MSL-WN), revealing only 16\% synset overlap between sign and spoken languages, confirming systematic differences exist but noting that ``the nature, frequency, and distribution of these differences remain unexplored.''

\paragraph{Research Gap.}
Previous work has shown that spoken and sign languages differ in how they resolve sense disambiguation of a word and a sign. However, it remains unknown how their senses correspond. While recent work has emphasized the need for linguistically-informed sign language processing models \citep{yin-etal-2021-including}, cross-modal ambiguity mapping remains largely unaddressed. Our work addresses this gap by presenting two methods to identify the type of sense correspondence between a word and its sign translation(s).
Furthermore, previous datasets in sign languages are limited in scope. For instance, PopSign ASL \citep{starner2023popsign} focuses on identifying homonyms in sign languages, but it is not of relevance to spoken languages. MSL-WN \citep{schulder2024signs} links sign languages to the multilingual WordNet, but it does not annotate correspondence between word uses and signs. Our dataset provides 1,404 manually annotated mappings between German word uses and DGS sign IDs across 32 words and 49 signs, with three types of sense correspondence.

\section{Our Dataset}
To investigate sense correspondence between German and DGS, we construct a novel dataset by combining complementary linguistic resources.

\subsection{Data Sources}

We combine two complementary linguistic resources: the German Word Usage Graphs (D-WUGs) \citep{schlechtweg-etal-2024-refwug} and the Digital Dictionary of German Sign Language (DW-DGS) \citep{langer-etal-2024-introducing}.

\paragraph{D-WUG} provides word uses with human-annotated semantic proximity judgments on a four-point scale \citep{schlechtweg-etal-2020-semeval}. Uses are compiled into weighted, undirected graphs where nodes represent individual uses and edge weights correspond to median semantic proximity judgments. Correlation Clustering infers sense groupings a posteriori \citep{Bansal2004}, preserving gradedness while identifying empirically grounded semantic structures. Our analysis draws from three German datasets: DWUG\_DE (10 matched words), DiscoWUG (13 matched words), and RefWUG (9 matched words), spanning two historical periods (1800--1899 and 1946--1990).

\paragraph{DW-DGS} provides corpus-validated sign senses 
through video recordings and micons (moving icons) representing signs visually \citep{langer-etal-2024-introducing}. Each unique sign receives an ID, making it easily distinguishable. The dictionary provides German translation equivalents and ``Erklärung'' (clarification, or the explanation of the sense). 

\paragraph{Video IDs Instead of Glosses.} 
Following the DGS's practice \cite{otte2022sign}, we use video IDs rather than sign glosses to mark distinct signs, as the glosses do not reliably capture homonymy and polysemy in DGS. This issue also applies to ASL (see two examples below):

\begin{tcolorbox}[
  title=\textbf{Example 1: FRECH (DGS)},
  colback=white,
  colframe=black,
  boxrule=0.8pt,
  arc=2mm,
  left=1pt,
  right=6pt,
  top=1pt,
  bottom=1pt
]
\textbf{Homonymous Overlap.} A single sign form corresponds to two completely unrelated meanings:
\begin{itemize}[leftmargin=9pt]
    \item Meaning 1: "frech" (cheeky/impudent) describing behavior
    \item Meaning 2: "USA" (West Berlin regional variant) referring to the country
\end{itemize}
Using a single gloss obscures these unrelated meanings sharing the same sign form.
\end{tcolorbox}

\begin{tcolorbox}[
  title=\textbf{Example 2: DEAF (ASL)},
  colback=white,
  colframe=black,
  boxrule=0.8pt,
  arc=2mm,
  left=1pt,
  right=6pt,
  top=1pt,
  bottom=1pt
]
\textbf{Movement and Location Variation.} Analysis of 1,618 tokens from seven U.S. 
regions reveals that grammatical function is the primary constraint on this variation 
\citep{bayley2000phonological}. Three phonologically distinct variants all receive 
the gloss "DEAF":
\begin{itemize}[leftmargin=9pt]
    \item Variant A: Citation Form (ear to chin, downward path)
    \item Variant B: Reversed (chin to ear, upward path)
    \item Variant C: Contact-Cheek (cheek only, no path movement)
\end{itemize}
\end{tcolorbox}

\subsection{Human Annotation}

We search DW-DGS for each unique word from the D-WUG dataset and identify 32 matching entries (split as 21/11 for ``train + val + test\_overlap'' and test\_no\_overlap sets). Human annotation focuses on mapping each word use within D-WUG to its corresponding DGS sign(s), based on the German translation equivalents and ``Erklärung'' (clarifications) in DW-DGS. We use DW-DGS entry ID numbers instead of glosses to mark signs, as multiple DGS signs may share the same gloss but associate with different DW-DGS entries. Our annotation labels three sense correspondence types:

\paragraph{Type 1 (One-to-Many).} A German word is polysemous or homonymous and corresponds to multiple DW-DGS signs, indicating that DGS distributes the word senses across several signs. 

\paragraph{Type 2 (Many-to-One).} Multiple German words correspond to a single polysemous or homonymous DW-DGS sign, suggesting that DW-DGS compresses several word senses into one sign form. 

\paragraph{Type 3 (One-to-One).} 
A German word and its DW-DGS sign translation exhibit parallel senses.

\paragraph{No Match.} Although the German word appears in both WUG and DGS, their senses do not match.

Our human annotation proceeds as follows: (1) extract all word uses from WUG for each target word, (2) collect all DW-DGS signs matching the word with their German translations and sense clarifications, (3) manually map each word use to appropriate sign ID(s), (4) manually assign a suitable  
correspondence type.
For example, ``Behandlung'' (treatment/handling) maps to two signs (IDs 637, 999) for medical versus processing contexts, yielding Type 1; "Abend" (evening) maps to one sign (ID 19) that also corresponds to "Nacht" (night), yielding Type 2.

Details of our dataset are outlined in Table~\ref{tab:data_stats}. In the development set (train+val+test\_overlap), type distribution of headwords has relatively balanced coverage: Type 1 (28.6\%), Type 2 (28.6\%), Type 3 (33.3\%), and No Match (9.5\%).

\begin{table}[t]
\footnotesize
\centering
\setlength{\tabcolsep}{2.5pt}
\begin{tabular}{l|c|cccc}
\toprule
\textbf{Split} & \textbf{Level} & \textbf{T1} & \textbf{T2} & \textbf{T3} & \textbf{NM} \\
\midrule
train + val + test\_overlap & mapping & 573 & 147 & 236 & 17 \\
test\_no\_overlap & mapping & 168 & 84 & 87 & 92 \\
\midrule
train + val + test\_overlap & word
 & 6 & 6 & 7 & 2 \\
test\_no\_overlap & word
 & 3 & 3 & 3 & 2 \\
\midrule
\multicolumn{5}{c}{\textbf{Total: 1,404 mappings from 32 words and 49 signs}} \\
\bottomrule
\end{tabular}
\caption{Dataset statistics showing balanced distribution across sense correspondence types, with Type 1 (one-to-many) being most common at the instance level (573 instances, 59.0\% of development set).}
\label{tab:data_stats}
\end{table}

\section{Our Approach}
We implement two computational methods to automatically identify sense correspondence types between German words and their DGS sign translations.

\paragraph{Exact Match (EM)} retrieves candidate DW-DGS signs by looking for lexical overlap between the two D-WUG entries (a headword and its word use) and the two DW-DGS entries (German translations and ``Erklärung'' of each sign).

When a match is found, the corresponding DW-DGS video entry ID is retrieved as a candidate. All candidates are then ranked by computing the semantic similarities between D-WUG and DW-DGS entries based on their embeddings. Note that EM uses semantic similarity only for ranking retrieved candidates, whereas SS uses it for both retrieval and ranking.

\paragraph{Semantic Similarity (SS)} encodes the same D-WUG entries and the same DW-DGS entries (as in EM) into embeddings by using SBERT \citep{sentence-transformers}. 
We use $q$ to denote the concatenated embedding of the D-WUG entries, while using $m$ to denote the concatenated embedding of the DW-DGS entries. We compute the cosine similarity between $q$ and $m$:

\begin{align}
    \text{sim}(q, m) = \frac{\mathbf{e}_q \cdot \mathbf{e}_m}{|\mathbf{e}_q| |\mathbf{e}_m|} = \cos(\theta)
\end{align}

The similarity score ranges from $-1$ to $1$, with values closer to $1$ indicating stronger semantic alignment. We evaluate four sentence embedding models: paraphrase-multilingual-MiniLM-L12-v2, all-MiniLM-L6-v2, German\_Semantic\_STS\_V2, and German-roberta-sentence-transformer-v2 \citep{reimers2019sentencebert,reimers-gurevych-2020-making}.

\paragraph{Sense correspondence type.} 
For each German word $w$, we collect the predicted DGS sign IDs, together with their semantic similarity scores that are previously denoted by $\text{sim}(q,m)$. We let $V_w$ be a set of DGS sign IDs predicted for word $w$. The sense Correspondence Type $\text{T}(w)$ is then assigned according to the following rules:
\begin{equation}
\text{T}(w) =
\begin{cases}
\text{No Match}, & \text{if } |V_w| = 0, \\
\text{Type 1}, & \text{if } |V_w| > 1, \\
\text{Type 2 }, &  \text{if } |V_w| = 1 \text{ AND } \text{sim} < \tau, \\
\text{Type 3}, & \text{if } |V_w| = 1 \text{ AND } \text{sim} \ge \tau
\end{cases}
\end{equation}
This applies to both SS and EM methods.

\section{Experimental Setup}
We evaluate our methods across multiple configurations to assess their effectiveness in identifying sense correspondence types.
\subsection{Data Splits}
For the development set, we partition the manually annotated data into three splits: (i) \textbf{the train split} (723 mappings), which serves as the candidate pool for retrieving potential sign matches; (ii) \textbf{the validation split} (100 mappings), used for hyperparameter tuning via grid search; and (iii) \textbf{the test\_overlap split} (150 mappings), used for model comparison and evaluation.

\subsection{Evaluation Setup} We conduct experiments in two setups: (i) \textbf{With vocabulary overlap}: the test\_overlap split contains words and signs that are present in the train set, enabling direct comparison between EM and SS methods and (ii) \textbf{Without vocabulary overlap}: the test\_no\_overlap set contains entirely different words and signs not in the train set, measuring whether SS can generalize to novel vocabulary. Since EM requires lexical matches, only SS is evaluated in the zero-overlap scenario.

\subsection{Hyperparameter optimization} The SS method has two hyperparameters: (i) similarity threshold $\tau$, determining minimum cosine similarity required and (ii) top-$k$, controlling how many high-scoring candidates are selected from each data source before merging. We optimise these hyperparameters via grid search on the validation split, totalling 12 configurations: $\tau \in \{0.65, 0.70, 0.75, 0.80\}$ and $k \in \{3,5,7\}$. Grid search is conducted separately for each embedding model.

\subsection{Evaluation Metrics}
Our metric is accuracy, defined as the proportion of cases where top-ranked predicted signs (including ties) match our human annotation. 
For No Match cases, a prediction is considered correct only if our methods return no prediction, i.e., no candidate exceeds the similarity threshold. 
Additionally, we evaluate the ranking quality beyond the top prediction by reporting Precision@$K$, which measures whether the correct sign appears within the top $K$ ranked candidates, where $K\in \{1,3,5,10\}$ \citep{jarvelin2002cumulated,Manning2008}.

Lastly, we break down our evaluation into individual sense correspondence types, examining performance gap across the three correspondence types (one-to-many, many-to-one, and one-to-one) as well as No Match cases. Additionally, we conduct an error analysis reporting error rate per type, and then look into the prediction agreement between EM and SS, for instance, analysing how often both methods succeed, how often both fail, how often only one method succeeds.

\subsection{Ablation Setups} We experiment with two ablation setups to evaluate the impact of different input components:
\begin{itemize}
    \item D-WUG entries: (i) \textbf{Full Context}, which combines a German word and its word use; (ii) \textbf{Word Only}, which uses the word only; and (iii) \textbf{Sentence Only}, which uses the word use only.
    \item DW-DGS entries: (i) \textbf{Base}, which uses both German translation equivalents (GT) and ``Erklärung'', and (ii) \textbf{GT Only}, which uses German translations only.
\end{itemize}

\section{Results}
We present the performance of our methods across different evaluation scenarios and correspondence types.

\subsection{Model Selection}

\begin{table}[t!]
    \footnotesize
    \centering
     \setlength{\tabcolsep}{3pt}
    \begin{tabular}{l | c c c c}
    \toprule
    \textbf{Model} & \textbf{Thr.} & \textbf{K} & \textbf{Acc.} & \textbf{Imp.} \\
    \midrule
    paraphrase-multi-MiniLM-L12-v2 & 0.65 & 3 & 78.57 & +8.33 \\
    all-MiniLM-L6-v2 & 0.70 & 3 & 73.81 & -1.19 \\
    German\_semantic\_sts\_v2 & 0.80 & 3 & 72.62 & +4.76 \\
    German-roberta-sent-v2 & 0.80 & 3 & 69.05 & -1.19 \\
    \bottomrule
    \end{tabular}
    \caption{Optimal hyperparameters for each embedding model on the validation split. The paraphrase-multilingual model achieves the highest accuracy (78.57\%) and shows the largest improvement over exact matching (+8.33 pp).}
    \label{tab:hyperparameter_optimization}
\end{table}

Table \ref{tab:hyperparameter_optimization} reports the hyperparameter configuration of each model. All the hyperparameters are tuned by using grid search on the validation data split. Then, we evaluate the models with these hyperparameters on the test\_overlap set. 
We find that 
paraphrase-multi-MiniLM-L12-v2 achieves the best accuracy 
(78.57\%) on the validation split, while all-MiniLM-L6-v2 achieves the best accuracy (88.52\%) on the test\_overlap split, with SS outperforming EM by a 17.21 percentage point. Thus, we use all-MiniLM-L6-v2 for the remaining analyses.

\begin{table}[t]
\footnotesize
\centering
\setlength{\tabcolsep}{3pt}
\begin{tabular}{l|ccc}
\toprule
\textbf{Model} & \textbf{EM} & \textbf{SS} & \textbf{Imp.} \\
\midrule
all-MiniLM-L6-v2 & 71.31 & 88.52 & 17.21 \\
German\_semantic\_sts\_v2 & 70.49 & 87.70 & 17.21 \\
paraphrase-multi-MiniLM-L12-v2 & 68.85 & 86.89 & 18.03 \\
German-roberta-sent-v2 & 71.31 & 84.43 & 13.11 \\
\bottomrule
\end{tabular}
\caption{Model performance on test\_overlap split demonstrates that all embedding models substantially outperform exact matching, with all-MiniLM-L6-v2 achieving the best accuracy (88.52\%, +17.21 pp improvement).}
\label{tab:model_comparison_validation}
\end{table}

\subsection{Overall Results}
Table~\ref{tab:best_model_overall} presents model accuracies on the test\_overlap split. We see that SS outperforms EM in all cases by a 17.21 percentage point, indicating approximately 24\% relative improvement.
Both SS and EM achieve a perfect score (100\%) in ``No Match'' cases. Thus, the improvement of SS stems from ``Match'' cases, where SS achieves 86.67\% accuracy compared to EM's 66.67\% (+20.0 pp).

\begin{table}[t]
\footnotesize
\centering
\begin{tabular}{l|ccc}
\toprule
\textbf{Category} & \textbf{EM} & \textbf{SS} & \textbf{Imp.} \\
\midrule
Overall (n=122) & 71.31 & 88.52 & 17.21 \\
Match (n=105) & 66.67 & 86.67 & 20.00 \\
No Match (n=17) & 100.0 & 100.0 & 0.0 \\
\bottomrule
\end{tabular}
\caption{Overall performance comparison shows semantic similarity (SS) outperforms exact matching (EM) by 17.21 percentage points, with the improvement stemming entirely from 'Match' cases where sense correspondence exists.}
\label{tab:best_model_overall}
\end{table}

\subsection{Type-Specific Results}
Table~\ref{tab:pathway_accuracy} shows how model accuracies vary across different sense correspondence types. 
Type 1 seems most challenging for EM (41.7\%) but benefits dramatically from SS (93.8\%, +52.1 pp), demonstrating that  SS relying on SBERT can identify one-to-many sense correspondence type in almost all cases. 
For Type 3, EM achieves a perfect score EM performance (100\%), while SS lags behind (88.9\%, -11.1 pp).
This suggests that when a German word and its sign translation has parallel senses, our SS is not so reliable, which may introduce noise and produce wrong matches. However, the advantage of EM over SS is likely due to a dataset artefact, namely the vocabulary overlap between data splits, which may not be generalisable to other datasets. For Type 2, both approaches are on par, indicating equal challenges for both.
For No Match, both approaches achieve a perfect score (100\%). 

\begin{table}[t]
\footnotesize
\centering
\begin{tabular}{l|c|ccc}
\toprule
\textbf{Type} & \textbf{n} & \textbf{EM} & \textbf{SS} & \textbf{Imp.} \\
\midrule
Type 1 & 48 & 41.7 & 93.8 & 52.1 \\
Type 2 & 21 & 66.7 & 66.7 & 0.0 \\
Type 3 & 36 & 100.0 & 88.9 & -11.1 \\
No Match & 17 & 100.0 & 100.0 & 0.0 \\
\bottomrule
\end{tabular}
\caption{Type-specific accuracy reveals dramatic differences: SS excels at Type 1 (one-to-many) with +52.1 pp improvement, EM performs best for Type 3 (one-to-one), while Type 2 (many-to-one) remains equally challenging for both methods.}
\label{tab:pathway_accuracy}
\end{table}

\subsection{Ranking Quality and Error Patterns}
In Table~\ref{tab:precision_k}, we find that SS improves from P@1 (79.5\%) to P@3 (88.5\%), while EM from 71.3\% to 91.8\%. Both plateau at $K$=3. In Table~\ref{tab:error_flow}, we see that SS successfully predicts 20.5\% of cases in which EM fails (25 instances), while it fails in 3.3\% of cases where EM succeeds (4 instances), yielding a net gain of +17.2 pp. In 68.0\% of cases, both methods succeed, whereas both fail in 8.2\% of cases (10 instances).

\begin{table}[t]
\footnotesize
\centering
\begin{tabular}{ll|c|cc}
\toprule
\textbf{Method} & \textbf{Category} & \textbf{n} & \textbf{P@1} & \textbf{P@3} \\
\midrule
\multirow{3}{*}{SS} & Overall & 122 & 79.5 & 88.5 \\
 & Match & 105 & 76.2 & 86.7 \\
 & No Match & 17 & 100.0 & 100.0 \\
\midrule
\multirow{3}{*}{EM} & Overall & 122 & 71.3 & 91.8 \\
 & Match & 105 & 66.7 & 90.5 \\
 & No Match & 17 & 100.0 & 100.0 \\
\bottomrule
\end{tabular}
\caption{Ranking quality analysis shows both methods plateau at P@3, with EM achieving slightly higher precision (91.8\%) than SS (88.5\%) when considering top-3 predictions, suggesting EM provides better candidate ranking despite lower top-1 accuracy.}
\label{tab:precision_k}
\end{table}

\begin{table}[t]
\footnotesize
\centering
\begin{tabular}{l|cc}
\toprule
\textbf{Pattern} & \textbf{Count} & \textbf{\%} \\
\midrule
Both Succeed & 83 & 68.0 \\
SS Success, EM Fail & 25 & 20.5 \\
EM Success, SS Fail & 4 & 3.3 \\
Both Fail & 10 & 8.2 \\
\bottomrule
\end{tabular}
\caption{Error pattern analysis reveals SS succeeds in 20.5\% of cases where EM fails, while failing in only 3.3\% of cases where EM succeeds, yielding a net gain of +17.2 pp and demonstrating complementary strengths.}
\label{tab:error_flow}
\end{table}

\subsection{Confusion Matrix}
We use a confusion matrix to report the agreement between ground-truth and model predictions (by SS) of sense correspondence types on the test\_overlap set (20 words) (see Tables~\ref{tab:pathway_confusion} and~\ref{tab:pathway_agreement}).
Overall agreement reaches 60\% (12/20 words with correct predictions of sense correspondence types). No Match achieves perfect agreement (100\%, 2/2), followed by Type 3 at 85.7\% (6/7). Type 1 achieves 66.7\% agreement (4/6). Our SS 
fails to identify Type 2 instances (0/5), never predicting Type 2, instead misclassifying those cases as No Match (2 words) or Type 3 (3 words).

\begin{table}[t]
\footnotesize
\centering
\setlength{\tabcolsep}{4pt}
\begin{tabular}{l|ccc}
\toprule
\textbf{Type} & \textbf{Total} & \textbf{Correct} & \textbf{Agr. (\%)} \\
\midrule
No Match & 2 & 2 & 100.0 \\
Type 1 & 6 & 4 & 66.7 \\
Type 2 & 5 & 0 & 0.0 \\
Type 3 & 7 & 6 & 85.7 \\
\midrule
Overall & 20 & 12 & 60.0 \\
\bottomrule
\end{tabular}
\caption{Agreement rates between ground truth and predictions show perfect performance for No Match (100\%), strong performance for Type 3 (85.7\%), moderate for Type 1 (66.7\%), but complete failure for Type 2 (0\%), resulting in 60\% overall agreement.}
\label{tab:pathway_agreement}
\end{table}

\subsection{Ablation Studies}

\begin{table}[t]
\footnotesize
\centering
\setlength{\tabcolsep}{4pt}
\begin{tabular}{l|ccc}
\toprule
\multicolumn{4}{c}{\textbf{D-WUG Entries Ablation}} \\
\midrule
\textbf{Input Mode} & \textbf{EM} & \textbf{SS} & \textbf{Imp.} \\
\midrule
Word Only & 72.95 & 91.80 & 18.85 \\
Full Context & 71.31 & 88.52 & 17.21 \\
Sentence Only & 73.77 & 88.52 & 14.75 \\
\midrule
\multicolumn{4}{c}{\textbf{DW-DGS Entries Ablation}} \\
\midrule
\textbf{Configuration} & \textbf{EM} & \textbf{SS} & \textbf{Imp.} \\
\midrule
Base (GT + ``Erklärung'') & 71.31 & 88.52 & 17.21 \\
GT Only & 71.31 & 88.52 & 17.21 \\
Difference & 0.00 & 0.00 & 0.00 \\
\bottomrule
\end{tabular}
\caption{Ablation studies show that (1) encoding the German word alone achieves the best SS accuracy (91.80\%), outperforming full context by +3.28 pp, suggesting word usage context introduces noise, and (2) including dictionary definitions provides no gain over German translations alone (all values in \%).}
\label{tab:ablation_combined}
\end{table}

Table~\ref{tab:ablation_combined} reports ablation results for both data sources. For D-WUG entries, ``Word Only'' achieves the best SS accuracy (91.80\%), outperforming ``Full Context'' by +3.28 pp, while ``Sentence Only'' yields the best EM performance (73.77\%). The underperformance of ``Full Context'' suggests that jointly encoding the word with its usage introduces noise that negatively affects predictions. For DW-DGS entries, including ``Erklärungen'' (dictionary definitions) yields no performance difference compared to using German translations alone for both methods, indicating that translation equivalents contain sufficient semantic information.

\subsection{Generalization (test\_no\_overlap)}

Our method strongly relies on vocabulary overlap between data splits. In Table \ref{tab:testnooverlap}, we find that model performance drops sharply from the overlap to the no-overlap setting (88.52\% to 17.59\%), suggesting that the overlap between the test\_overlap split and the train split is crucial. Without such overlap, our SS
based on word uses only cannot address sense correspondence between a word and its sign translation. 

Our ablation studies support this finding: Using ``Word Only'' performs better than using ``Full Context'' that combines word and its uses (91.80\% vs. 88.52\%)
as shown on Table \ref{tab:ablation_combined}, indicating the importance of vocabulary. Although incorporating word uses from WUG datasets does not improve performance here, such contextual information may be beneficial for other tasks across spoken and sign languages. 

\begin{table}[t]
\footnotesize
\centering
\setlength{\tabcolsep}{3pt}
\begin{tabular}{l|ccc}
\toprule
\textbf{Metric} & \textbf{W/Ovlp} & \textbf{W/o Ovlp} & \textbf{Gap} \\
\midrule
Overall Acc. (\%) & 88.52 & 17.59 & -70.93 \\
\midrule
Match (n) & 105 & 431 & --- \\
Match Acc. (\%) & 86.67 & 0.00 & -86.67 \\
\midrule
No Match (n) & 17 & 92 & --- \\
No Match Acc. (\%) & 100.00 & 100.00 & 0.00 \\
\midrule
Type 1 Agr. (\%) & 66.7 & 0.0 & -66.7 \\
Type 3 Agr. (\%) & 85.7 & 0.0 & -85.7 \\
Overall Agr. (\%) & 60.0 & 18.2 & -41.8 \\
\bottomrule
\end{tabular}
\caption{test\_no\_overlap analysis reveals severe performance degradation without vocabulary overlap (88.52\% to 17.59\%, -70.93 pp), demonstrating that the model's success critically depends on seeing the same words during training rather than generalizing to semantic patterns.}
\label{tab:testnooverlap}
\end{table}

\subsection{Parts-of-speech Analysis}

We also conduct a parts-of-speech analysis to examine whether different word classes exhibit distinct sense correspondence patterns. Appendix~\ref{sec:pos_analysis} provides detailed results. Table~\ref{tab:word_type_dev_gt} shows that verbs predominantly map to Type 1 (3 of 6), nouns favor Types 2 and 3 combined (8 of 10), and adjectives split evenly between Types 1 and 3. Table~\ref{tab:word_type_test_comparison} reveals that in the zero-overlap scenario, our SS model predicts ``No Match'' for 9 of 11 words despite only 2 being genuine ``No Match'' cases, further confirming poor test\_no\_overlap without vocabulary overlap.

\section{Discussion}
Our findings reveal systematic patterns in how senses are organized across spoken and sign language modalities. 
Our analysis investigates how senses are organised differently across spoken and sign language modalities. 
The relatively balanced distribution of headwords across sense correspondence types (Type 1: 28.6\%, Type 2: 28.6\%, Type 3: 33.3\%) demonstrates that no single correspondence type dominates sense organisation across spoken and sign languages, while confirming the differences between the two modalities \citep{schulder2024signs,taub2001}. 

Sense correspondence between spoken and sign languages is dominated by Type 1 (50.0\%, 13 out of 26 instances) at the level of word–to–sign pairs, where the senses of a polysemous or homonymous spoken word are distributed across multiple signs. This aligns with findings by \citet{kristoffersen2010making} that sign language uses the visual-gestural modality to encode fine-grained senses across multiple signs. 

Performance gaps across correspondence types highlight the limitations of our methods. For Type 1 cases, SS outperforms EM, indicating that the SBERT that SS relies on can address correspondence between a polysemous word and multiple sign candidates, precisely the scenario where EM fails. For Type 3, where words and signs exhibit parallel senses, SBERT-based SS can sometimes overgeneralise by matching semantically similar but incorrect sign candidates. Type 2 is equally challenging for both EM and SS, as both methods struggle to distinguish whether a single sign has one sense or multiple senses. 

Our approaches are evaluated using both accuracy (accounting for ties) and P@K. The 9 percentage point difference between accuracy (88.52\%; see Table~\ref{tab:model_comparison_validation}) and P@1 (79.5\%; see Table~\ref{tab:precision_k}) indicates that in approximately 9\% of cases on the test\_overlap set, SS identifies the ground-truth sign but does not rank it first.
Our analysis also shows that SS is effective at retrieving semantically related correspondence that EM fails to detect, while EM provides better ranking performance than SS when the correct candidates are retrieved.

\section{Conclusion}

This work provides the first structured analysis of sense correspondence types between spoken German and German Sign Language. Our work analyses how lexical senses align across spoken and sign languages. We manually annotated sense correspondence between German words and DGS signs, and presented computational methods to identify the types of sense correspondence. We found that the senses of German words and their sign translations are organised differently across three correspondence types.

We contribute a dataset of 1,404 manually annotated word use–to–sign ID mappings derived from 32 words 49 signs, establishing the first resource of its own kind.   
Our computational evaluation identifies which correspondence patterns are identifiable: SS is effective when a German word corresponds to multiple signs (+52.1 pp), EM performs best when a word and its sign translation have parallel senses, while both methods struggle when multiple German words correspond to a single sign. 
These findings reveal which correspondence patterns current approaches can identify:
Type 1 (one-to-many) correspondences are highly identifiable through SS, Type 3 (one-to-one) benefits from EM, while Type 2 (many-to-one) remains challenging for both methods.
Our findings advance prior work. While \citet{schulder2024signs} demonstrate systematic differences exist through 16\% synset overlap, noting the nature, and distribution of these differences remain unexplored, our analysis clarifies how these differences manifest across sense correspondence types and which cases are more difficult to identify using our approaches. While fine-grained sign language sense disambiguation remains challenging \cite{grimm2024signlanguagesensedisambiguation}, our results identify specific scenarios in which our methods succeed and others where they fall short.

Future work should enlarge our dataset, expand to other languages, include annotations conducted by native German–DGS bilingual speakers, explore multimodal embeddings that integrate sign videos, investigate dialectal variation using DW-DGS, and study semantic change across spoken and sign languages.

\section*{Limitations}

Our annotations were not conducted by native German–DGS bilingual speakers. We restricted each word–sign pair to a single sense correspondence type. 
Our dataset only contains 32 words and 49 signs, which is small; however, our focus is 1,404 human-annotated mappings from word usages to signs.
Our findings are limited to the German-DGS language pair and may not generalise to other languages. Finally, our approach relies heavily on vocabulary overlap; without such overlap, accuracy drops largely from 88.52\% to 17.59\%.

\bibliography{custom}
\clearpage
\section*{APPENDIX}
\appendix

\section{More Tables}

\begin{table}[htbp]
\begin{threeparttable}
\centering
\caption[Hyperparameter Optimization Results - Full]{Optimal hyperparameter configurations for each model (complete version).}
\label{tab:hyperparameter_optimization_full}
\footnotesize
\begin{tabular}{|p{3.2cm}|c|c|c|c|}
\hline
\textbf{Model} & \textbf{Thr.} & \textbf{K} & \textbf{Acc.} & \textbf{Imp.} \\
\hline
paraphrase-multi-lingual-MiniLM-L12-v2 & 0.65 & 3 & 78.57 & +8.33 \\
\hline
all-MiniLM-L6-v2 & 0.70 & 3 & 73.81 & -1.19 \\
\hline
German\_Sem-antic\_STS\_V2 & 0.80 & 3 & 72.62 & +4.76 \\
\hline
german-roberta-sentence-trans-former-v2 & 0.80 & 3 & 69.05 & -1.19 \\
\hline
\end{tabular}
\begin{tablenotes}
\footnotesize
\item Models ranked by optimization semantic accuracy. Thr.=Threshold, K=Top-K, Acc.=Semantic Accuracy (\%), Imp.=Improvement (\%). Batch size=64 for all models. Improvement refers to SS performance over EM on the validation set.
\end{tablenotes}
\end{threeparttable}
\end{table}

\begin{table}[htbp]
\begin{threeparttable}
\centering
\caption[Precision@K Analysis - Full]{Precision@K results on test split (complete version).}
\label{tab:precision_k_full}
\footnotesize
\renewcommand{\arraystretch}{1.2}
\begin{tabular}{|l|l|c|c|c|c|}
\hline
\textbf{Meth.} & \textbf{Category} & \textbf{n} & \textbf{P@1} & \textbf{P@3} & \textbf{P@5} \\
\hline
\multirow{3}{*}{SS} & Overall & 122 & 79.5 & 88.5 & 88.5 \\
 & Match & 105 & 76.2 & 86.7 & 86.7 \\
 & No-Match & 17 & 100.0 & 100.0 & 100.0 \\
\hline
\multirow{3}{*}{EM} & Overall & 122 & 71.3 & 91.8 & 91.8 \\
 & Match & 105 & 66.7 & 90.5 & 90.5 \\
 & No-Match & 17 & 100.0 & 100.0 & 100.0 \\
\hline
\end{tabular}
\begin{tablenotes}
\footnotesize
\item P@K = percentage of cases where correct sign appears within top K predictions. P@10 values identical to P@5 (omitted for space). For No Match cases, P@K measures correct abstention.
\end{tablenotes}
\end{threeparttable}
\end{table}

\begin{table}[htbp]
\begin{threeparttable}
\centering
\caption[Input Component Ablation - Full]{Input component ablation results on test split (complete version).}
\label{tab:input_ablation_full}
\footnotesize
\renewcommand{\arraystretch}{1.2}
\begin{tabular}{|l|c|c|c|c|}
\hline
\textbf{Input Mode} & \textbf{EM} & \textbf{SS} & \textbf{Imp.} & \textbf{Conf.} \\
\hline
Word Only & 72.95 & 91.80 & 18.85 & 0.830 \\
Full Context & 71.31 & 88.52 & 17.21 & 0.746 \\
Sentence Only & 73.77 & 88.52 & 14.75 & 0.746 \\
\hline
\end{tabular}
\begin{tablenotes}
\footnotesize
\item EM=EM Accuracy (\%), SS=Semantic Accuracy (\%), Imp.=Improvement (\%), Conf.=Average Confidence. All configurations use best model (all-MiniLM-L6-v2) with optimized hyperparameters (threshold=0.70, top\_k=3). Full Context represents baseline configuration.
\end{tablenotes}
\end{threeparttable}
\end{table}
\clearpage

\begin{table}[t]
\footnotesize
\centering
\begin{tabular}{l|cccc}
\toprule
\textbf{GT $\downarrow$ / Pred $\rightarrow$} & \textbf{NM} & \textbf{T1} & \textbf{T3} & \textbf{Total} \\
\midrule
No Match & 2 & 0 & 0 & 2 \\
Type 1 & 1 & 4 & 1 & 6 \\
Type 2 & 2 & 0 & 3 & 5 \\
Type 3 & 1 & 0 & 6 & 7 \\
\midrule
Total & 6 & 4 & 10 & 20 \\
\bottomrule
\end{tabular}
\caption{Confusion matrix on test set (20 words) shows the model never predicts Type 2, instead misclassifying all 5 Type 2 cases as either No Match (2) or Type 3 (3), indicating fundamental difficulty distinguishing polysemous signs.}
\label{tab:pathway_confusion}
\end{table}

\section{Computational Typology Classification Visualization}

\label{app:confusion_matrix}

\begin{figure}[h]
    \centering
    \includegraphics[width=0.5\textwidth]{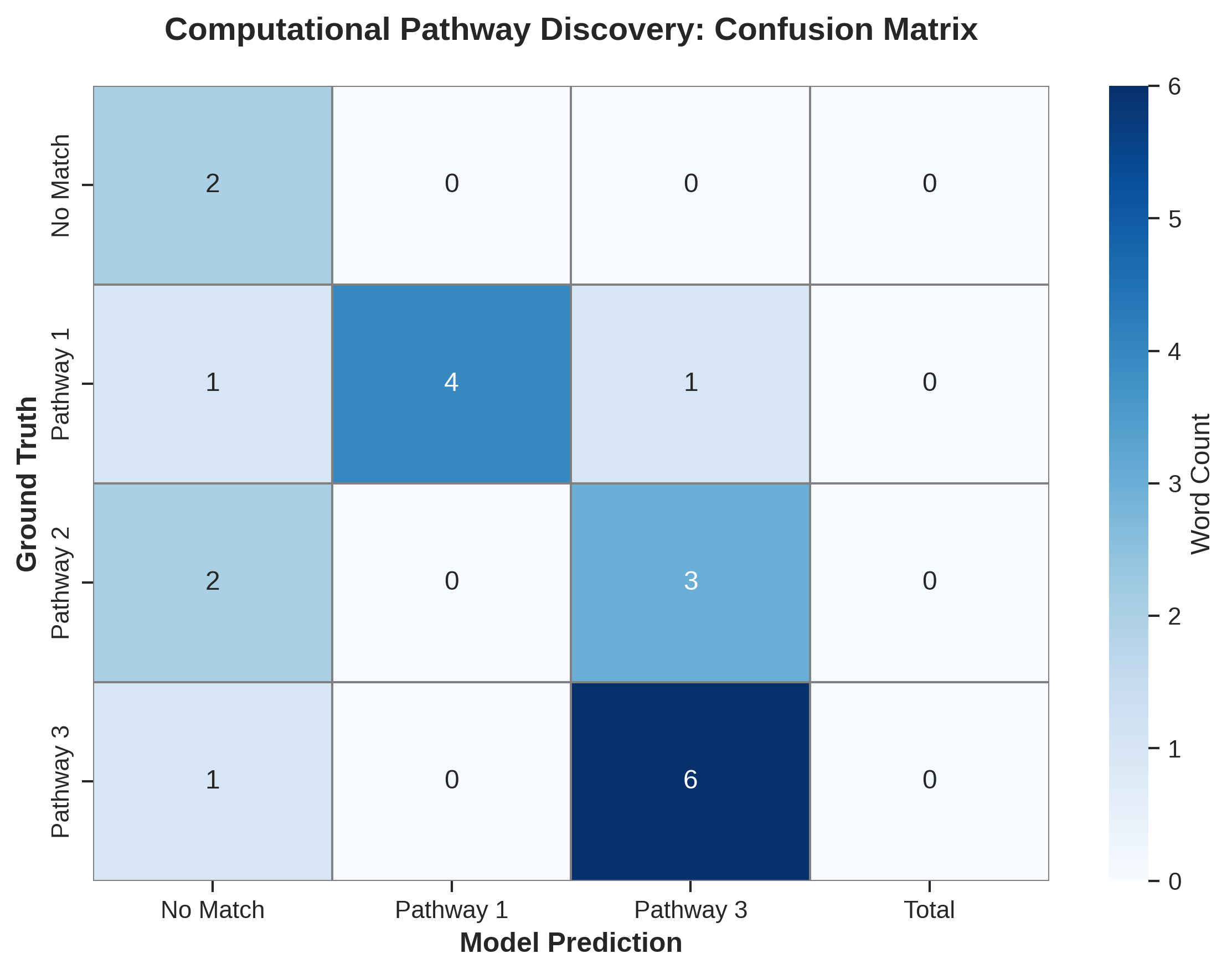}
    \caption{Confusion matrix visualizing agreement between model predictions and human annotations across 20 test words, clearly showing the model's complete inability to identify Type 2 (many-to-one) correspondence, with all 5 Type 2 cases misclassified as either No Match or Type 3.}
    \label{fig:confusion}
\end{figure}

Figure~\ref{fig:confusion} shows agreement between computational typology discovery and human annotations across 20 test words. The model achieves strong agreement for No Match (100\%, 2/2) and Type 3 (85.7\%, 6/7), moderate agreement for Type 1 (66.7\%, 4/6), but completely fails to identify Type 2 (0/5). The model never predicts Type 2, instead misclassifying these cases as No Match (2) or Type 3 (3), suggesting fundamental difficulty in distinguishing whether a single sign represents one coherent meaning or multiple unrelated meanings.

\section{Parts-of-Speech Analysis}
\label{sec:pos_analysis}

Table~\ref{tab:word_type_dev_gt} presents the distribution of sense correspondence types across different parts of speech in the development set, revealing systematic patterns in how word classes map to correspondence types.

\begin{table}[ht]
\footnotesize
\centering
\begin{tabular}{lll}
\textbf{POS} & \textbf{Word} & \textbf{GT Type} \\
\toprule
Verb & ausbilden & Type 2 \\
Verb & bemerken & Type 3 \\
Verb & eintreten & Type 3 \\
Verb & erlauben & Type 1 \\
Verb & freigelassen & Type 1 \\
Verb & helfen & Type 1 \\
\midrule
Noun & Museum & Type 2 \\
Noun & Mauer & Type 1 \\
Noun & Vorbereitung & Type 2 \\
Noun & Westen & Type 3 \\
Noun & Behandlung & Type 1 \\
Noun & Entscheidung & Type 3 \\
Noun & Frechheit & Type 2 \\
Noun & Mut & Type 3 \\
Noun & Seminar & Type 3 \\
Noun & Tier & Type 2 \\
\midrule
Adjective & englisch & Type 1 \\
Adjective & finnisch & Type 3 \\
\bottomrule
\end{tabular}
\caption{Parts-of-speech distribution in development set shows verbs predominantly map to Type 1 (one-to-many, 3 of 6), nouns favor Types 2 and 3 combined (8 of 10), and adjectives split evenly, suggesting word class influences sense correspondence patterns.}
\label{tab:word_type_dev_gt}
\end{table}

Table~\ref{tab:word_type_test_comparison} compares ground-truth and model predictions in the zero-vocabulary-overlap scenario, demonstrating the model's failure to generalize beyond memorized vocabulary.

\begin{table}[ht]
\centering
\footnotesize
\begin{tabular}{llll}
\toprule
\textbf{POS} & \textbf{Word} & \textbf{GT Type} & \textbf{Model Type} \\
\midrule
Noun & Anstellung & Type 1 & No Match \\
Adjective & billig & Type 3 & No Match \\
Noun & Zufall & Type 3 & No Match \\
Noun & Presse & Type 3 & No Match \\
Verb & packen & Type 1 & No Match \\
Verb & anpflanzen & No Match & No Match \\
Verb & niederschlagen & No Match & No Match \\
Verb & abbauen & Type 2 & Type 3 \\
Verb & artikulieren & Type 3 & No Match \\
Noun & Schmiere & Type 2 & No Match \\
Noun & Titel & Type 2 & No Match \\
\bottomrule
\end{tabular}
\caption{Model predictions in zero-vocabulary-overlap scenario show the model incorrectly predicts ``No Match'' for 9 of 11 words despite only 2 genuine cases, confirming poor test\_no\_overlap and over-reliance on lexical memorization.}
\label{tab:word_type_test_comparison}
\end{table}

\end{document}